\documentclass[]{relaxed_system_lab}

\usepackage[utf8]{inputenc}
\usepackage[T1]{fontenc}
\usepackage{charter}
\usepackage{varwidth}
\usepackage{wrapfig}
\usepackage{fontawesome5}

\usepackage{multirow}
\usepackage{booktabs}       
\usepackage[table]{xcolor}  
\usepackage{array}
\usepackage{tabularx}

\newcolumntype{Y}{>{\centering\arraybackslash}X}
\newcolumntype{Z}{>{\raggedright\arraybackslash}X}

\usepackage{amsmath, amssymb, amsfonts, bm} 
\usepackage{graphicx}          

\usepackage{xspace}            
\usepackage{enumitem}          

\usepackage{makecell}          

\usepackage{float}
\usepackage{soul}

\usepackage{natbib}
\usepackage{latexsym}

\usepackage{url}
\usepackage{amssymb}
\usepackage[utf8]{inputenc}
\usepackage{microtype}
\usepackage{booktabs}
\usepackage{pifont} 
\usepackage{multirow}
\usepackage{makecell}
\usepackage{xspace}
\usepackage{color}
\usepackage{xcolor}
\usepackage{colortbl}
\usepackage{adjustbox}
\usepackage{hyperref} 
\usepackage[edges]{forest}
\usepackage{tikz} 
\usepackage{caption}
\usepackage{amsfonts}

\hypersetup{
    colorlinks,
    linkcolor={blue!80!black},
    citecolor={blue!80!black},
}
\tikzset{
    root/.style =             {align=center, text width=1cm, rounded corners=3pt, line width=0.3mm, fill=gray!10, draw=gray!80, font=\small},
    demographic/.style =         {align=center, text width=1.8cm, rounded corners=3pt, line width=0.3mm, fill=blue!10, draw=blue!80, font=\footnotesize},
    demographic_work/.style =    {align=center, text width=10cm, rounded corners=3pt, line width=0.3mm, fill=blue!10, draw=blue!0, font=\footnotesize},
    character/.style =         {align=center, text width=1.8cm, rounded corners=3pt, line width=0.3mm, fill=red!10, draw=red!80, font=\footnotesize},
    character_work/.style =    {align=center, text width=10cm, rounded corners=3pt, line width=0.3mm, fill=red!10, draw=red!0, font=\footnotesize},
    personalization/.style =           {align=center, text width=1.8cm, rounded corners=3pt, line width=0.3mm, fill=cyan!10, draw=cyan!80, font=\footnotesize},
    personalization_work/.style =      {align=center, text width=10cm, rounded corners=3pt, line width=0.3mm, fill=cyan!10, draw=cyan!0, font=\footnotesize},
    risk/.style =         {align=center, text width=1.8cm, rounded corners=3pt, line width=0.3mm, fill=orange!10, draw=orange!80, font=\footnotesize},
    risk_work/.style =    {align=center, text width=10cm, rounded corners=3pt, line width=0.3mm, fill=orange!10, draw=orange!0, font=\footnotesize},
}

\usepackage{CJK}

\DeclareTextFontCommand{\textbf}{\bfseries}
\usepackage{needspace}
\usepackage{flafter}
\renewcommand{\arraystretch}{1.16}
\newcommand{\rqfinding}[2]{%
  \FloatBarrier\par\addvspace{5pt}\Needspace{4\baselineskip}%
  \begingroup
  \setlength{\fboxsep}{7pt}\setlength{\fboxrule}{0.5pt}%
  \noindent\fcolorbox{oursAccent!65}{oursRow}{%
    \parbox{\dimexpr\linewidth-2\fboxsep-2\fboxrule\relax}{%
      \small\textbf{#1 key observation.} #2}}%
  \par\endgroup\addvspace{5pt}%
}

\definecolor{lightgreen}{RGB}{144,238,144}
\definecolor{darkcyan}{HTML}{008B8B}

\newcommand{\sys}{\textsc{GameASG-Bench}\xspace}

\definecolor{gsmDark}{HTML}{315B73}
\definecolor{humanDark}{HTML}{39706D}
\definecolor{mathDark}{HTML}{665A7A}
\definecolor{logiqaDark}{HTML}{8A653B}

\definecolor{gsmLight}{HTML}{E8F1F6}
\definecolor{humanLight}{HTML}{E7F2F0}
\definecolor{mathLight}{HTML}{F0EDF5}
\definecolor{logiqaLight}{HTML}{F7F0E7}

\definecolor{oursRow}{HTML}{EAF3FF}
\definecolor{alternateRow}{HTML}{F7F8FA}
\definecolor{oursAccent}{HTML}{2563A6}

\definecolor{headerGray}{HTML}{F1F2F3}
\definecolor{subheaderGray}{HTML}{F8F8F8}

\title{\sys: Benchmarking Autonomous Software Generation for Game Development}

\author{Xiuhui Zhang$^{1,2}$ , Yi Chen$^1$, Shusheng Xu$^1$, Fan Li$^1$, Huan Wang$^1$,
Tongkai Yang$^1$, Binhang Yuan$^{1,3}$}

\affiliation{$^1$Ant Group, $^2$Beihang University, $^3$HKUST}

\abstract{
Autonomous software generation (ASG) aims to turn human requirements into executable applications, but delivering these applications does not necessarily establish that their interacting components satisfy the specified behavioral requirements. We introduce \sys, a benchmark that makes behavioral testability part of the generation task for game development. Our design declares an evaluation interface specification before generation, fixing legal starting scenarios, player-level actions, stable snapshots, rejection behavior, and invariants while leaving private implementations open. Concretely, we include: (\underline{i}) static L1 checks that assess source-level compliance; and (\underline{ii}) browser-executed L2 checks that combine semantic observations with real input and runtime evidence. We implement this protocol as 47 browser-native game-generation tasks spanning 12 primary genres and both 2D and 3D interaction, each with executable checks and an independently verified reference implementation. Our experiments answer four key questions about end-to-end agent performance, tool access and nominal turn budget, reasoning effort, and harness choice. Across nine agent stacks, the highest observed mean L2 check pass rate is 93.2\%, yet the highest observed strict task success rate, requiring all L1 and applicable L2 prerequisite and core requirement checks, is only 55.3\% (26/47 tasks). For DeepSeek-V4-Flash, full tool access and larger nominal turn budgets yield more strict task successes, while the strict task success rate is not monotonic in reasoning effort. Both tested harnesses achieve 18 strict task successes, but only ten tasks succeed under both. These results expose task-level compliance gaps that high average check pass rates actually obscure.
}

\begin{document}
\maketitle

\begin{center}
\href{https://github.com/areal-project/GameASG-Bench}{%
  \faGithub\,\texttt{\:Code: github.com/areal-project/GameASG-Bench}}
\vspace{0.5em}
\end{center}

\section{Introduction}
\label{sec:intro}

Autonomous software generation (ASG) aims to translate human requirements into complete, executable software artifacts with minimal human intervention, extending the scope of agentic coding from local implementation tasks to end-to-end development~\citep{liu2026e2edev}. Modern coding agents embed large language models (LLMs) in loops that inspect files, edit code, invoke tools, execute programs, and revise failed implementations~\citep{yang2024sweagent}. Success on bounded programming problems or repository-level edits, however, does not establish that an agent can deliver an application whose interacting components collectively satisfy the original requirements~\citep{liu2026e2edev,zhang2026webgamebench}. We therefore ask: \textit{how can we systematically evaluate whether frontier LLMs, operating through contemporary coding-agent loops, can autonomously generate a complete, executable application whose runtime behavior satisfies the specified behavioral requirements?} We study browser-native game generation as a bounded and behavior-dense proxy for autonomous requirement-to-artifact generation. The resulting benchmark measures compliance with specified gameplay and evaluation requirements.

Component-level correctness provides an incomplete view of ASG~\citep{liu2026e2edev,liu2026mage}. A generated program may parse and contain plausible mechanics, yet remain unusable because its controls, shared state, rendering, and user-interaction logic do not work together~\citep{zhang2026webgamebench,chen2026gamexpert}. Complete games make these integration requirements observable: even a compact game must coordinate input handling, spatial behavior, state transitions, scoring or resource changes, failure conditions, and restart within one executable artifact~\citep{zhang2026webgamebench}. Evaluating these interactions helps distinguish successful delivery from partial implementation and identifies specific behavioral failures~\citep{jia2026gamegenverifier}. By itself, an aggregate ``playable'' judgment offers limited evidence about which specified behavior failed or why, motivating checks that attribute failures to individual requirements~\citep{jia2026gamegenverifier,xu2026videovibe}.

These integration requirements complicate both generation and evaluation~\citep{luo2026gamecraft,jia2026gamegenverifier}. During \emph{generation}, the agent must translate a natural-language specification into consistent state transitions, interfaces, rendering logic, and player interactions~\citep{luo2026gamecraft}. For example, a movement routine can be locally plausible yet update the wrong entity, operate during a locked phase, consume the wrong resource, or fail to affect the rendered game. During \emph{evaluation}, successful compilation does not establish that the requested mechanics exist, while open-ended playtesting can miss conditions that require long or precise interaction sequences~\citep{liu2026mage,jia2026gamegenverifier}. Controlled state access makes such conditions easier to exercise, and bounded verification can combine visual traces with assertions over runtime state~\citep{jia2026gamegenverifier}. In our setting, evaluator-facing state is implemented by the generated artifact and must be checked against browser behavior; For example, in Section~\ref{sec:threats}, we illustrate how scenario preparation, semantic observations, and real input can fail to compose. We therefore design an evaluator around reproducible legal starting conditions, player-level actions, and evidence connecting state changes to browser input and execution, while preserving freedom in the artifact's private implementation (Section~\ref{sec:system}).

Recent benchmarks address some aspects of this problem. Function-level benchmarks test bounded programs~\citep{hendrycks2021apps,liu2023evalplus}; repository benchmarks evaluate changes to existing software~\citep{jimenez2024swebench}; and interactive-agent benchmarks measure operation of already-built environments~\citep{zhou2024webarena}. Recent game benchmarks move closer to ASG. GameDevBench~\citep{chi2026gamedevbench} evaluates complex modifications to Godot projects; WebGameBench~\citep{zhang2026webgamebench} evaluates complete browser-native games through real-browser interaction; GameCraft-Bench~\citep{luo2026gamecraft} evaluates end-to-end Godot generation using replay and multimodal evidence; and GameGen-Verifier~\citep{jia2026gamegenverifier} decomposes specifications into bounded precondition-interaction-postcondition checks supported by runtime state injection. 

However, how to give executable checks consistent access to heterogeneous generated artifacts remains an underexplored design question. Thus, we propose \sys, which explores a complementary approach: the task author specifies behavioral testability \emph{before generation} through a common evaluation interface, legal scenario definitions, stable semantic observations, and explicit invariants. The coding agent implements the specified interface together with the game, allowing fixed checks to combine semantic observations with real browser input and independent runtime evidence. Concretely, we make the following key contributions:

\textbf{\underline{Contribution 1.}}
We design \sys to evaluate requirement-to-artifact generation through a pre-declared evaluation interface specification (Section~\ref{sec:system}). Legal starting scenarios, player-level actions, stable snapshots, rejection behavior, and invariants provide fixed checks with consistent access to different private implementations. L1 checks assess source-level compliance; L2 checks exercise the game in headless Chromium using semantic observations, real browser input, and runtime evidence. We distinguish these evidence layers from requirement priorities and define strict task success as successful delivery and evaluation with every L1 check and every applicable L2 prerequisite and core requirement check passing.

\textbf{\underline{Contribution 2.}}
We implement this design as 47 browser-native game-generation tasks spanning 12 primary genres and both 2D and 3D interaction (Section~\ref{sec:benchmark-construction}). Each task packages a generation prompt, a gameplay design requirement, an evaluation interface specification, and executable checks, with a self-contained \texttt{index.html} as the required artifact. Task authors translate gameplay requirements into scenario, action, observation, and invariant definitions before generation. Quality control combines independent verification of reference implementations with automated acceptance testing, connecting the shared evaluation protocol to a concrete task corpus.

\textbf{\underline{Contribution 3.}}
We conduct experiments to answer four research questions about end-to-end performance, tool access and nominal turn budget, reasoning effort, and harness choice (Section~\ref{sec:evaluation}). Across nine agent stacks, the highest observed mean L2 check pass rate is 93.2\%, yet the highest observed strict task success rate is 55.3\% (26/47 tasks), showing that high average check pass rates can mask unmet task requirements. For DeepSeek-V4-Flash, full tool access yields 18 strict task successes versus 6--9 with limited tool access; the strict task success rate over all 47 planned tasks rises from 7/47 to 18/47 as the nominal turn budget increases from 30 to 120 as more tasks complete evaluation. High reasoning effort yields 19 strict task successes versus 18 at maximum reasoning effort while using 26.9\% fewer reasoning tokens. Both tested harnesses achieve 18 strict task successes, but only ten tasks succeed under both.

\section{Benchmark Design and Evaluation Protocol}
\label{sec:system}

The core design of \sys is to make testability part of the generation task. The human developer defines an evaluation interface specification before generation, and the coding agent implements the specified interface together with the game. Fixed checks can then prepare scenarios and exercise player-level actions across different implementations. This section describes the shared task structure, evaluation interface, evidence layers, and scoring rule; Section~\ref{sec:benchmark-construction} explains how that design is instantiated in the benchmark corpus. The distinguishing design choice is how responsibilities are divided among the human developer, coding agent, and evaluator, building on established ideas of behavioral interface specifications, controllability, and behavioral assertions~\citep{meyer1992applying,hatcliff2012behavioral,freedman1991testability,barr2015oracle}.

\subsection{Task Definition and Overview}
\label{sec:task-definition}
\label{sec:system-execution}

Each \sys task requires a coding agent to produce a self-contained browser-native game in \texttt{index.html}. The game must satisfy both its player-visible requirements and the evaluation interface specification that makes those requirements testable. Three documents define the task (Table~\ref{tab:task-package}): the \emph{generation prompt} specifies the workspace and delivery protocol; the \emph{gameplay design requirement} describes the playable loop, mechanics, feedback, and completion conditions; and the \emph{evaluation interface specification} defines the evaluation interface and its task-specific semantics.

\begin{table}[htbp]
    \centering
    \small
    \caption{Task documents and their roles. All three are available to the coding agent; only the generation prompt is passed directly to the harness.}
    \label{tab:task-package}
    \begin{tabularx}{\linewidth}{@{}p{0.30\linewidth}lZ@{}}
        \toprule
        \textbf{Document} & \textbf{File} & \textbf{Content} \\
        \midrule
        Generation prompt & \texttt{target.md} & Workspace, delivery protocol, and concise gameplay brief. \\
        \addlinespace[3pt]
        Gameplay design requirement & \texttt{game-spec.md} & Player-visible mechanics and the minimum playable loop. \\
        \addlinespace[3pt]
        Evaluation interface specification & \texttt{tdd.md} & Scenario setup, action and observation semantics, rejection behavior, and invariants. \\
        \bottomrule
    \end{tabularx}
\end{table}

The human developer prepares these documents and executable checks before generation. Each attempt starts in a clean workspace with a fresh agent session. The harness passes \texttt{target.md} directly to the coding agent, which consults the two specifications and implements the game and evaluation interface together. The evaluator checks delivery, applies fixed L1 and L2 checks, and records itemized outcomes and strict task success. This division specifies the controllable actions and observable behavior while leaving the private implementation open. Harness invocations, delivery preflight, and access boundaries appear in Section~\ref{sec:eval-setup} and Appendix~\ref{app:implementation}.

\subsection{Evaluation Interface Specification}
\label{sec:system-principles}

Each evaluation interface specification defines four methods exposed through \texttt{window.\_\_gameTest}: \texttt{reset}, \texttt{loadScenario}, \texttt{input}, and \texttt{getSnapshot}. Scenario names, actions, and snapshot fields are task-specific. The interface provides two complementary capabilities.

\textbf{Reproducible starting conditions.}
The \texttt{reset} operation restores the initial state and clears transient effects. \texttt{loadScenario} sets a documented legal state reachable through ordinary play, including rare or late-game preconditions. Scenario setup may adjust resources and positions to establish the required starting state, but it must not directly produce the outcome being tested. A near-terminal scenario, for example, must still require play to cause victory or defeat. This constraint preserves the role of the tested action while bounding the cost of reaching its precondition.

\textbf{Actions and observations with stable meanings.}
The \texttt{input} operation performs player-level actions such as ordering a unit or selecting a target. \texttt{getSnapshot} returns a JSON-serializable summary of the relevant game state. The evaluation interface specification fixes the meanings of actions and observations, including rejection behavior and invariants, while permitting different internal representations. The agent need not expose its private object graph or reproduce a prescribed code layout.

All four methods operate on the underlying state that drives visible gameplay. State changes made through \texttt{reset}, \texttt{loadScenario}, and \texttt{input} are reflected in the game, while \texttt{getSnapshot} reports the corresponding state. This connection allows the evaluator to prepare scenarios, perform actions, and inspect outcomes within the actual game.

\subsection{Layered Evaluation and Behavioral Evidence}
\label{sec:system-layers}

Evaluation separates source-level compliance from executed gameplay behavior. L1 inspects the delivered artifact; L2 exercises it in a browser. The layers run in that order, but an L1 failure does not suppress L2, so a report can distinguish a missing source declaration from a runtime failure.

\begin{itemize}[leftmargin=*,nosep]
    \item \textit{L1: source-level compliance.} A shared runner inspects the HTML and its inline or directly linked local scripts. The task's \texttt{checks.json} declares structural and syntax checks, regular-expression assertions, and anti-pattern rules. The runner records an outcome and diagnostic hints for each check. These checks establish evidence of required source patterns or interface declarations, not gameplay correctness: a regular-expression match, for example, may occur in a comment or string.
    \vspace{0.5em}
    
    \item \textit{L2: controlled behavioral execution.} The evaluator serves the submission locally and executes \texttt{checks.js} in headless Chromium. Behavioral checks use a prepare-act-observe sequence: they can establish a documented scenario, invoke a semantic action or send real browser input, and compare observed state changes with expected outcomes and invariants. Prerequisite checks may only verify startup and the required interface. Checks run sequentially, each in a newly created browser page that loads the submitted game. After a check completes or times out, the runner closes its page before starting the next check. A shared browser hook records animation frames, input listeners, and Canvas or WebGL activity; checks can also inspect snapshots, rendered output, and runtime errors. Snapshots report task-specific game state, while real input and browser observations help assess whether the reported changes are reflected in visible gameplay. The combination of evidence sources depends on the individual check; not every check uses every signal. Each check returns \texttt{PASS}, \texttt{FAIL}, or \texttt{NOT\_APPLICABLE}, together with elapsed time and diagnostic detail.
\end{itemize}

\subsection{Requirement Priorities and Scoring}
\label{sec:system-scoring}

We use L1 and L2 to specify evidence layers, and P0, P1, and P2 to specify requirement priorities. Prerequisite checks (P0) cover launch, evaluation-interface availability, and minimum runtime requirements. Core requirement checks (P1) cover required mechanics, interactions, invariants, and supporting interface requirements. Extended capability checks (P2) cover additional mechanics and experience completeness. Priorities are assigned during test authoring. L2 checks marked \texttt{NOT\_APPLICABLE} are excluded when calculating applicable-check pass rates. The current runner does not enforce a restriction against returning this outcome for required P1 checks.

For each planned task $g$, let $d_g,e_g\in\{0,1\}$ indicate valid delivery and completed evaluation, respectively. Let $\mathcal{L}_g$ contain every L1 check, and let $\mathcal{B}_g$ contain the applicable L2 P0/P1 checks. For each check $c$, $p_{g,c}=1$ if the check passes and $0$ otherwise. Strict task success $s_g$ and the primary strict task success rate over the set of planned tasks $\mathcal{G}$ are
\begin{equation}
\label{eq:strict-success}
s_g=d_g e_g\!\prod_{c\in\mathcal{L}_g\cup\mathcal{B}_g}p_{g,c},
\qquad
\operatorname{SR}(\mathcal{G})=\frac{1}{|\mathcal{G}|}\sum_{g\in\mathcal{G}}s_g.
\end{equation}
Thus, strict task success requires valid delivery, completed evaluation, and passing all L1 checks and all applicable L2 P0/P1 checks. The primary strict task success rate is calculated over all planned tasks, including generation and evaluation failures, and is reported as a percentage. When reported, conditional strict task success rates use only tasks that complete evaluation as the denominator.

To characterize partial compliance, we also report check pass rates over tasks that complete evaluation. Mean L1 and overall L2 check pass rates are calculated by averaging per-task check pass rates. For each L2 priority, the pass rate is computed by pooling applicable checks across evaluated tasks and dividing the number passed by the total. These metrics complement strict task success by showing how many checks pass even when a task does not satisfy every required check.

\FloatBarrier

\section{Benchmark Construction}
\label{sec:benchmark-construction}

We instantiate the design in Section~\ref{sec:system} through task selection, specification and test authoring, and quality control, yielding 47 tasks, each with executable checks and a reference implementation. This section describes that construction process and the resulting corpus; Section~\ref{sec:evaluation} reports reference acceptance and agent performance.

\subsection{Scope and Task Selection}
\label{sec:task-selection}

Each selected task must support a complete playable loop: the player's input changes the game state and produces observable progress or a terminal outcome, and the game supports restarting. Tasks must fit the self-contained \texttt{index.html} delivery protocol. We exclude concepts requiring a backend, user accounts, external databases, paid or private assets, unbounded multiplayer infrastructure, or behavior that cannot be reached and observed within bounded browser execution. This scope concentrates evaluation on integration among controls, state transitions, rendering, and gameplay logic.

\subsection{Specification and Test Authoring}
\label{sec:data-preparation}
\label{sec:test-authoring}

For each selected game concept, human developers write a gameplay design requirement specifying the objective, controls, entity roles, state transitions, scoring or resource effects, player feedback, terminal conditions, and restart behavior. They then prepare the generation prompt by combining the gameplay brief with the delivery instructions, following the document structure defined in Section~\ref{sec:task-definition}.

Human developers then define task-specific scenarios, actions, and snapshot fields in the evaluation interface specification. For each tested behavior, they specify the starting conditions, the player-level action, the expected observable outcomes, and the invariants that must hold. For invalid actions, they specify the expected rejection behavior and which parts of the game state must remain unchanged. Scenario setup and interface behavior follow the constraints described in Section~\ref{sec:system-principles}.

L1 checks are specified in \texttt{checks.json}, and L2 behavioral checks are implemented in \texttt{checks.js}. Human developers assign requirement priorities according to Section~\ref{sec:system-scoring}.

\subsection{Quality Control}
\label{sec:quality-control}

Quality control combines independent verification of reference implementations with automated acceptance testing. For each task, the reference implementation is independently verified through manual inspection against the gameplay design requirements and evaluation interface specification. Verification covers real user interactions, core state transitions, termination and restart, and consistency between the evaluation interface and visible gameplay. The authored L1 and L2 checks are then run on the verified implementation. Reference acceptance requires passing all L1 checks and all applicable L2 P0/P1 checks; L2 P2 outcomes are recorded separately. This procedure uses independently verified implementations as positive controls to assess whether the test suite accepts games that satisfy the core requirements. Section~\ref{sec:eval-setup} reports the acceptance results.

\subsection{Corpus Composition}
\label{sec:task-categorization}

\begin{figure}[htbp]
    \centering
    \captionsetup{format=plain,labelsep=space,font={small,rm},labelfont=bf,textfont=normalfont,justification=raggedright,singlelinecheck=false,skip=7pt}
    \captionsetup[subfigure]{format=plain,labelsep=space,labelfont=bf,textfont=normalfont,font={small,rm},justification=centering,singlelinecheck=true,skip=4pt}
    \begin{subfigure}[t]{0.49\linewidth}
        \vspace{0pt}
        \centering
        \includegraphics[width=\linewidth]{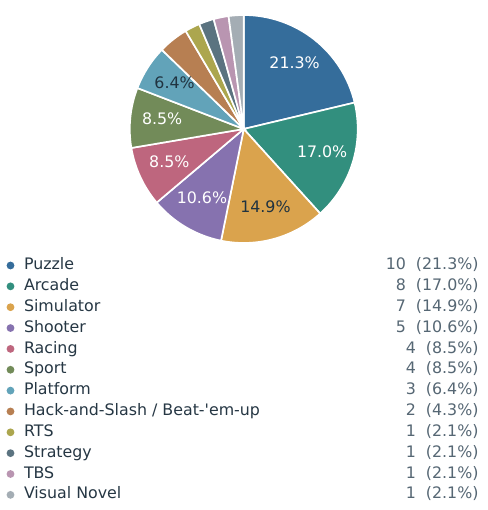}
        \caption{Genre}
        \label{fig:taxonomy-genre}
    \end{subfigure}\hfill
    \begin{minipage}[t]{0.49\linewidth}
        \vspace{0pt}
        \begin{subfigure}[t]{\linewidth}
            \centering
            \includegraphics[width=\linewidth]{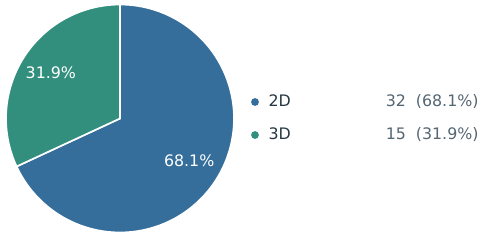}
            \caption{Dimension}
            \label{fig:taxonomy-dimension}
        \end{subfigure}
        \par\vspace{4pt}
        \begin{subfigure}[t]{\linewidth}
            \centering
            \includegraphics[width=\linewidth]{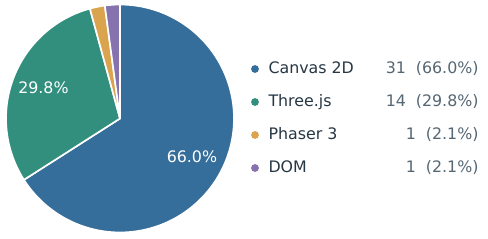}
            \caption{Reference technology}
            \label{fig:taxonomy-technology}
        \end{subfigure}
    \end{minipage}
    \caption{Composition of the 47 tasks. Each panel covers all 47 tasks; legend entries give counts and percentages. RTS and TBS denote real-time and turn-based strategy, respectively. Reference technology identifies the rendering technology used by each reference implementation.}
    \label{fig:task-taxonomy}
\end{figure}

The benchmark comprises 47 tasks spanning 12 primary genres, including 32 tasks with 2D environments and 15 with 3D environments. Task diversity is described along three axes: genre, dimension, and reference technology (Figure~\ref{fig:task-taxonomy}). \emph{Genre} describes the repeated core action and objective; \emph{dimension} describes the playable environment's spatial relations; and \emph{technology} describes the rendering route of the reference implementation. Most reference implementations use Canvas 2D or Three.js. Technology labels characterize the corpus and do not constrain compliant generated artifacts.

Across the 47 tasks, the test suite contains 336 L1 checks and 885 L2 checks, including 534 L2 core requirement checks (P1). Table~\ref{tab:construction-statistics} summarizes the checks using the layers and priorities defined in Section~\ref{sec:system}.

\begin{table}[htbp]
    \centering
    \small
    \caption{Check counts for the 47 tasks used in all reported experiments. Layer and priority describe different properties of a check.}
    \label{tab:construction-statistics}
    \begin{tabular*}{\linewidth}{@{\extracolsep{\fill}}lrlr@{}}
        \toprule
        \multicolumn{2}{l}{\textbf{L1: source-level checks}} & \multicolumn{2}{l}{\textbf{L2: behavioral checks}} \\
        \cmidrule(lr){1-2}\cmidrule(l){3-4}
        Tool checks & 43 & Prerequisite checks (P0) & 102 \\
        Regular-expression checks & 288 & Core requirement checks (P1) & 534 \\
        Anti-pattern checks & 5 & Extended capability checks (P2) & 249 \\
        \midrule
        \textbf{Total} & \textbf{336} & \textbf{Total} & \textbf{885} \\
        \bottomrule
    \end{tabular*}
\end{table}

\FloatBarrier

\section{Experimental Evaluation}
\label{sec:evaluation}

The experimental evaluation is designed to answer four key research questions: 
\begin{itemize} [leftmargin=*,nosep]
    \item (\textbf{\underline{RQ1}}) \textit{How well do contemporary agent stacks satisfy end-to-end task requirements?} 

    \vspace{0.5em}
    \item(\textbf{\underline{RQ2}}) \textit{How do tool access and nominal turn budget affect task performance?}

    \vspace{0.5em}
    \item(\textbf{\underline{RQ3}}) \textit{How does reasoning effort affect task performance and cost?}

    \vspace{0.5em}
    \item (\textbf{\underline{RQ4}}) \textit{How does harness choice affect task performance for a fixed model and reasoning-effort setting?}
\end{itemize}

We first describe the shared experimental setup and report acceptance results for the reference implementations. We then address the four research questions in turn, followed by diagnostic case analyses.

\subsection{Experimental Setup}
\label{sec:eval-setup}

\textbf{Tasks and configurations.}
All experiments use the same 47 tasks and task packages. Each task-configuration pair is run once from a clean workspace. Evaluation uses headless Chromium with a fixed $1280\times800$ viewport.
An \emph{agent stack} is a model paired with its coding harness: Claude Code 2.1.206 or Codex CLI 0.153.4. RQ1 uses each stack's maximum reasoning-effort setting and full tool access. RQ2--RQ4 reuse DeepSeek-V4-Flash with Claude Code, maximum reasoning effort, full tool access, and a nominal turn budget of 120 as the common baseline. Each subsection specifies the changes to this baseline.

\textbf{Evaluation metrics.}
The primary metric is the strict task success rate over all 47 planned tasks, as defined in Equation~\ref{eq:strict-success}. We also report check pass rates over tasks that complete evaluation. Mean L1 and overall L2 check pass rates are calculated by averaging per-task check pass rates; L2 check pass rates for each priority are computed by pooling applicable checks within each priority and dividing the number passed by the total. We additionally report the number of tasks that deliver artifacts and complete evaluation, along with available measurements of artifact size, token use, and monetary cost.

\textbf{Reference-implementation validation.}
Following the verification procedure in Section~\ref{sec:quality-control}, we evaluated one independently verified core-compliant reference implementation for each of the 47 tasks. All 47 passed all L1 checks and all applicable L2 P0/P1 checks.

\subsection{RQ1: End-to-End Agent Performance}
\label{sec:rq-end-to-end}

RQ1 asks how reliably contemporary coding agents satisfy the complete task requirements. We compare nine agent stacks using the same task packages and evaluation protocol. All nine stacks deliver artifacts and complete evaluation on all 47 tasks. Table~\ref{tab:end-to-end} reports strict task success and check pass rates, and Table~\ref{tab:generation-resources} reports resource use.

\begin{table}[htbp]
    \centering
    \small
    \setlength{\tabcolsep}{4pt}
    \caption{End-to-end results at maximum reasoning effort. Strict task success is reported as the number of successful tasks out of 47 and the corresponding rate (\%); all other entries are check pass rates (\%). L1 and overall L2 entries are mean per-task check pass rates; L2 P0--P2 entries pool applicable checks within each priority. Bold denotes the highest observed value in each column.}
    \label{tab:end-to-end}
    \begin{tabular*}{\textwidth}{@{\extracolsep{\fill}}llrrrrrr@{}}
        \toprule
        \textbf{Model} & \textbf{Harness}
        & \makecell[r]{\textbf{Strict task}\\\textbf{success}}
        & \textbf{L1} & \multicolumn{4}{c}{\textbf{L2}} \\
        \cmidrule(l){5-8}
        & & & & \textbf{Mean} & \textbf{P0} & \textbf{P1} & \textbf{P2} \\
        \midrule
        GPT-6-Astra       & Codex CLI   & \textbf{26/47 (55.3)} & 98.0 & \textbf{93.2} & 99.0 & \textbf{92.7} & \textbf{92.4} \\
        Claude-Opus-5     & Claude Code & 24/47 (51.1) & \textbf{99.6} & 90.4 & 91.2 & 89.9 & 91.1 \\
        GPT-5.6-Sol       & Codex CLI   & 21/47 (44.7) & 98.9 & 91.3 & \textbf{100.0} & 89.7 & 91.9 \\
        DeepSeek-V4-Flash & Claude Code & 18/47 (38.3) & 99.4 & 89.3 & 98.0 & 86.7 & 90.2 \\
        DeepSeek-V4-Pro   & Claude Code & 15/47 (31.9) & \textbf{99.6} & 85.2 & 85.3 & 83.9 & 88.9 \\
        Kimi-K3           & Claude Code & 15/47 (31.9) & 97.7 & 88.5 & 98.0 & 86.3 & 89.7 \\
        GLM-5.2           & Claude Code & 11/47 (23.4) & 98.8 & 84.6 & 91.2 & 83.3 & 83.0 \\
        Hunyuan-3         & Claude Code & 10/47 (21.3) & 98.3 & 81.7 & 96.1 & 78.2 & 82.3 \\
        MiniMax-M3        & Claude Code & 7/47 (14.9)  & 98.3 & 70.6 & 91.2 & 64.5 & 75.4 \\
        \bottomrule
    \end{tabular*}
\end{table}

\begin{table}[htbp]
    \centering
    \small
    \setlength{\tabcolsep}{4pt}
    \caption{Mean artifact size and reported resource use. Artifact sizes cover all 47 tasks per stack. Input tokens combine the reported input, cache-read, and cache-write categories.}
    \label{tab:generation-resources}
    \begin{tabular*}{\textwidth}{@{\extracolsep{\fill}}llrrrr@{}}
        \toprule
        \textbf{Model} & \textbf{Harness}
        & \makecell[r]{\textbf{Artifact size}\\(KB)}
        & \makecell[r]{\textbf{Input tokens}\\(k)}
        & \makecell[r]{\textbf{Output tokens}\\(k)}
        & \makecell[r]{\textbf{Monetary}\\\textbf{cost} (USD)} \\
        \midrule
        GPT-6-Astra       & Codex CLI   & 78.82 & 896.1    & 55.5  & 4.4445 \\
        Claude-Opus-5     & Claude Code & 68.52 & 12,632.6 & 142.4 & 11.8500 \\
        GPT-5.6-Sol       & Codex CLI   & 64.67 & 1,382.5  & 45.7  & 1.7867 \\
        DeepSeek-V4-Flash & Claude Code & 85.21 & 13,038.4 & 203.6 & 0.2461 \\
        DeepSeek-V4-Pro   & Claude Code & 90.34 & 12,944.3 & 202.9 & 0.7431 \\
        Kimi-K3           & Claude Code & 57.51 & 1,263.9  & 72.3  & 2.5518 \\
        GLM-5.2           & Claude Code & 60.33 & 1,067.1  & 47.0  & 0.6184 \\
        Hunyuan-3         & Claude Code & 52.88 & 2,474.5  & 88.9  & 0.1779 \\
        MiniMax-M3        & Claude Code & 81.76 & 4,918.9  & 71.1  & 0.4069 \\
        \bottomrule
    \end{tabular*}
\end{table}

We identify three key observations:

\textit{High source-level pass rates coexist with incomplete task compliance.}
Mean L1 check pass rates range from 97.7\% to 99.6\%, whereas strict task success rates range from 14.9\% to 55.3\% (Table~\ref{tab:end-to-end}). GPT-6-Astra with Codex CLI achieves the highest observed strict task success rate, at 26/47 (55.3\%). Only two stacks succeed on more than half of the tasks, and even the stack with the highest strict task success rate fails to satisfy all required checks on 21 tasks. High average source-level check pass rates therefore do not translate into equally high strict task success rates.

\textit{Average check pass rates and strict task success capture different outcomes.}
GPT-6-Astra achieves the highest observed mean L2 check pass rate (93.2\%) and L2 P1 and P2 check pass rates (92.7\% and 92.4\%, respectively), yet its strict task success rate is 55.3\%; only 40 of its 47 artifacts pass every L1 check. Rankings can also differ: GPT-5.6-Sol has a higher mean L2 check pass rate than Claude-Opus-5 (91.3\% versus 90.4\%) but fewer strict task successes (21 versus 24). These results support reporting both check pass rates and strict task success: even a single failed required check prevents an artifact with otherwise high check pass rates from achieving strict task success.

\textit{Larger artifacts and higher token usage do not consistently correspond to more strict task successes.}
GPT-6-Astra records 55.5k mean output tokens and achieves 26 strict task successes, compared with approximately 203k output tokens for each DeepSeek variant and 18 and 15 strict task successes for Flash and Pro, respectively. MiniMax-M3 produces larger artifacts than Claude-Opus-5 (81.76 versus 68.52~KB) but achieves 7 rather than 24 strict task successes. DeepSeek-V4-Pro produces the largest artifacts (90.34~KB) and achieves 15 strict task successes. These contrasts show that larger artifacts or higher reported token usage do not consistently correspond to more complete task compliance.

\rqfinding{RQ1}{Across nine stacks, mean L1 check pass rates range from 97.7\% to 99.6\%, but strict task success rates range from 14.9\% to 55.3\%. High average check pass rates do not ensure that an artifact satisfies every required check.}

\FloatBarrier

\subsection{RQ2: Effect of Tool Access and Nominal Turn Budget}
\label{sec:rq-tools}

RQ2 examines DeepSeek-V4-Flash under different levels of tool access and nominal turn budgets. Both comparisons reuse the Claude Code baseline with full tool access, maximum reasoning effort, and a nominal turn budget of 120. The tool comparison covers four configurations: no tools (serialized API), file read/write only, files + syntax checking, and full tool access. The no-tools configuration uses serialized API input instead of the file-based harness workflow. The budget comparison uses full tool access with nominal turn budgets of 30, 60, and 120.

\begin{table}[htbp]
    \centering
    \small
    \setlength{\tabcolsep}{4pt}
    \caption{Tool configurations for DeepSeek-V4-Flash. All 47 tasks complete evaluation in each condition. Strict task success is reported as the number of successful tasks out of 47 and the corresponding rate (\%); check pass rates are percentages. Syntax checking permits Bash for \texttt{node -{}-check}, but no browser. The configuration with full tool access uses the nominal 120-turn baseline.}
    \label{tab:tool-ablation}
    \begin{tabular*}{\textwidth}{@{\extracolsep{\fill}}lrrrrr@{}}
        \toprule
        \textbf{Tool setting} & \makecell[r]{\textbf{Strict task}\\\textbf{success}}
        & \textbf{L1} & \textbf{L2 P0} & \textbf{L2 P1} & \textbf{L2 P2} \\
        \midrule
        No tools (serialized API) & 7/47 (14.9)  & 98.5 & 83.3 & 65.3 & 80.1 \\
        File read/write only     & 6/47 (12.8)  & 98.8 & 87.3 & 66.2 & 73.4 \\
        Files + syntax checking  & 9/47 (19.1)  & 98.2 & 86.3 & 63.4 & 69.3 \\
        Full tool access         & \textbf{18/47 (38.3)} & \textbf{99.4} & \textbf{98.0} & \textbf{86.7} & \textbf{90.2} \\
        \bottomrule
    \end{tabular*}
\end{table}

\textit{Full tool access achieves the most strict task successes and the highest check pass rates.}
Full tool access achieves 18 strict task successes, compared with 6--9 under restricted tool configurations, including 7 in the no-tools condition (Table~\ref{tab:tool-ablation}). It also achieves the highest P1 check pass rate, at 86.7\%, while L1 check pass rates remain close to 99\% across configurations. Adding syntax checking to file read/write access increases the number of strict task successes from 6 to 9, while the P1 check pass rate decreases from 66.2\% to 63.4\%. Thus, improvements in the number of strict task successes and check pass rates do not consistently coincide across the restricted configurations.
Appendix~\ref{app:repair-traces} presents examples of tool-assisted testing and repair during generation.

\begin{table}[htbp]
    \centering
    \small
    \setlength{\tabcolsep}{4pt}
    \caption{Nominal turn-budget comparison for DeepSeek-V4-Flash with full tool access. The primary strict task success rate uses all 47 planned tasks as the denominator; conditional strict task success rates and check pass rates describe only the evaluated subset. Counts are followed by percentages in parentheses; check pass rates are percentages.}
    \label{tab:turn-budget}
    \begin{tabular*}{\textwidth}{@{\extracolsep{\fill}}rrrrrrrr@{}}
        \toprule
        \makecell[r]{\textbf{Nominal turn}\\\textbf{budget}}
        & \makecell[r]{\textbf{Evaluated}\\\textbf{/ planned}}
        & \multicolumn{2}{c}{\textbf{Strict task success}}
        & \textbf{L1} & \textbf{L2 P0} & \textbf{L2 P1} & \textbf{L2 P2} \\
        \cmidrule(lr){3-4}
        & & \textbf{All planned} & \textbf{Conditional} & & & & \\
        \midrule
        30  & 10/47 & 7/47 (14.9)  & 7/10 (70.0)  & 100.0 & 100.0 & 94.0 & 97.6 \\
        60  & 32/47 & 13/47 (27.7) & 13/32 (40.6) & 99.7  & 88.6  & 81.6 & 88.5 \\
        120 & 47/47 & \textbf{18/47 (38.3)} & 18/47 (38.3) & 99.4 & 98.0 & 86.7 & 90.2 \\
        \bottomrule
    \end{tabular*}
\end{table}

\paragraph{Larger nominal turn budgets increase evaluation completion and strict task success rates.}
With nominal turn budgets of 30, 60, and 120, respectively, 10, 32, and 47 tasks complete evaluation, including 7, 13, and 18 strict task successes (Table~\ref{tab:turn-budget}). Over all 47 planned tasks, the strict task success rate therefore increases from 14.9\% to 27.7\% and 38.3\%. Among the 10, 32, and 47 evaluated tasks, the corresponding conditional strict task success rates are 70.0\%, 40.6\%, and 38.3\%, respectively.
Claude Code stops execution when the configured turn limit is reached. With smaller nominal turn budgets, more runs terminate before completing generation and delivery, reducing the number of tasks that reach evaluation. Appendix~\ref{app:budget-cases} provides examples of where these runs stop during verification and delivery.

\rqfinding{RQ2}{Full tool access achieves 18 strict task successes out of 47 tasks, compared with 6--9 under restricted tool configurations. Increasing the nominal turn budget from 30 to 120 raises the strict task success rate from 14.9\% to 38.3\%, while the number of tasks completing evaluation increases from 10 to 47.}

\FloatBarrier
\subsection{RQ3: Effect of Reasoning Effort}
\label{sec:rq-reasoning}

To explore RQ3, we fix DeepSeek-V4-Flash, Claude Code, full tool access, and a nominal turn budget of 120, and compare low, high, and maximum reasoning effort on the same 47 tasks. We measure computational cost by reasoning-token use. Table~\ref{tab:reasoning-ablation} reports strict task success, check pass rates, and mean reasoning-token use.

\begin{table}[htbp]
    \centering
    \small
    \setlength{\tabcolsep}{4pt}
    \caption{Reasoning-effort comparison for DeepSeek-V4-Flash. Strict task success is reported as the number of successful tasks out of 47 and the corresponding rate (\%); check pass rates are percentages. Reasoning-token counts are averaged across tasks. Bold denotes the highest observed number of strict task successes or check pass rate.}
    \label{tab:reasoning-ablation}
    \begin{tabular*}{\textwidth}{@{\extracolsep{\fill}}lrrrrrr@{}}
        \toprule
        \textbf{Effort} & \makecell[r]{\textbf{Strict task}\\\textbf{success}}
        & \textbf{L1} & \textbf{L2 P0} & \textbf{L2 P1} & \textbf{L2 P2}
        & \makecell[r]{\textbf{Reasoning}\\\textbf{tokens}} \\
        \midrule
        Low  & 7/47 (14.9)           & 98.3 & 93.1 & 71.7 & 80.2 & 37,699 \\
        High & \textbf{19/47 (40.4)} & \textbf{99.4} & 97.1 & 82.7 & 88.9 & 138,232 \\
        Max  & 18/47 (38.3)          & \textbf{99.4} & \textbf{98.0} & \textbf{86.7} & \textbf{90.2} & 189,210 \\
        \bottomrule
    \end{tabular*}
\end{table}

\textit{Core check pass rates increase, but the strict task success rate is not monotonic.}
The strict task success rate rises from 14.9\% (7/47 tasks) at low effort to 40.4\% (19/47) at high effort, then decreases to 38.3\% (18/47) at maximum effort. P1 check pass rates increase from 71.7\% to 82.7\% and 86.7\%, a low-to-maximum gain of 15.0 percentage points, the largest among the reported check pass rates. L1 changes from 98.3\% to 99.4\%, and P0 and P2 also increase. The principal observed improvement is therefore in core runtime requirements. Maximum effort achieves the highest P0, P1, and P2 check pass rates, while high effort achieves the most strict task successes.

\textit{High effort achieves more strict task successes with fewer reasoning tokens than maximum effort.}
Mean reasoning-token use increases from 37,699 at low effort to 138,232 at high effort and 189,210 at maximum effort. High effort uses 3.67 times as many reasoning tokens as low effort, alongside an increase from 7 to 19 strict task successes. Compared with maximum effort, high effort uses 26.9\% fewer reasoning tokens while achieving one more strict task success (19 versus 18). This identifies a useful trade-off in the recorded runs.

\rqfinding{RQ3}{High effort achieves the most strict task successes (19/47) with 26.9\% fewer reasoning tokens than maximum effort. Maximum effort achieves the highest P0, P1, and P2 check pass rates but yields 18/47 strict task successes.}

\FloatBarrier
\subsection{RQ4: Effect of Harness Choice}
\label{sec:rq-harness}

To answer RQ4, we run DeepSeek-V4-Flash at maximum reasoning effort through Claude Code 2.1.206 and Codex CLI 0.153.4, using the same 47 task packages, evaluation protocol, and full tool access. We compare the harnesses as complete execution environments, including their system instructions, context management, tool schemas, command execution, and endpoint protocols.

\begin{table}[htbp]
    \centering
    \small
    \setlength{\tabcolsep}{4pt}
    \caption{Harness comparison for DeepSeek-V4-Flash at maximum reasoning effort. Strict task success is reported as the number of successful tasks out of 47 and the corresponding rate (\%); check pass rates are percentages. Bold denotes the highest observed value in each column.}
    \label{tab:harness-ablation}
    \begin{tabular*}{\textwidth}{@{\extracolsep{\fill}}lrrrrr@{}}
        \toprule
        \textbf{Harness} & \makecell[r]{\textbf{Strict task}\\\textbf{success}}
        & \textbf{L1} & \textbf{L2 P0} & \textbf{L2 P1} & \textbf{L2 P2} \\
        \midrule
        Claude Code & \textbf{18/47 (38.3)} & \textbf{99.4} & \textbf{98.0} & 86.7 & 90.2 \\
        Codex CLI   & \textbf{18/47 (38.3)} & 99.2 & 96.1 & \textbf{87.1} & \textbf{92.2} \\
        \bottomrule
    \end{tabular*}
\end{table}

\textit{Equal numbers of strict task successes conceal different task outcomes.}
Both harnesses achieve strict task success on 18/47 tasks (38.3\%), but only ten tasks succeed under both. Eight succeed only under Claude Code, eight only under Codex CLI, and 21 under neither. The same aggregate count thus arises from different successful task sets. Check pass rates also differ across the two harnesses (Table~\ref{tab:harness-ablation}): Claude Code has higher L1 (99.4\% versus 99.2\%) and P0 (98.0\% versus 96.1\%) pass rates, whereas Codex CLI has higher P1 (87.1\% versus 86.7\%) and P2 (92.2\% versus 90.2\%) pass rates.

\textit{Resource profiles differ despite equal numbers of strict task successes.}
With the same number of strict task successes, Codex CLI produces smaller artifacts on average than Claude Code (69.5 versus 85.2~KB, an 18.4\% reduction) and records lower mean output-token use (112.8k versus 203.6k, a 44.6\% reduction).

\textit{Matched traces illustrate different repair sequences.}
The paired DeepSeek-V4-Flash \textit{Sortie} traces show both harnesses using tests to guide repairs. In Codex CLI, real-input testing exposes a coordinate offset after evaluation-interface self-tests pass. In Claude Code, self-tests reveal item states incorrectly preserved across reset, followed by edits to animation and audio initialization. Both final artifacts achieve strict task success. Appendix~\ref{app:repair-traces} details these repair sequences.

\rqfinding{RQ4}{Both harnesses achieve 18 strict task successes out of 47 tasks, with only ten tasks achieving strict task success under both. Equal numbers of strict task successes coexist with different successful task sets, check pass rates, and resource use.}

\subsection{Diagnostic Analysis}
\label{sec:threats}

The preceding experiments report strict task success and check pass rates. We now examine concrete failures to show how real-input checks and scenario-based behavioral checks identify violations of task requirements (Table~\ref{tab:diagnostic-evidence}). These cases connect the evaluator's design to the errors detected in generated artifacts. Appendix~\ref{app:failure-traces} provides the check-level evidence and supplementary failure analysis.

\begin{table}[htbp]
    \centering
    \small
    \renewcommand{\arraystretch}{1.12}
    \setlength{\tabcolsep}{5pt}
    \caption{Diagnostic evidence linking evaluation checks to observed failures. The GPT-6-Astra case uses Codex CLI; the DeepSeek-V4-Flash case uses Claude Code with maximum reasoning effort and full tool access.}
    \label{tab:diagnostic-evidence}
    \begin{tabularx}{\textwidth}{@{}p{0.22\textwidth}ZZ@{}}
        \toprule
        \textbf{Artifact / stack} & \textbf{Observed evidence} & \textbf{Role of the checks} \\
        \midrule
        \textit{Diner Dasher}\newline GPT-6-Astra
        & L1 and L2 P0/P2 pass; L2 P1 mouse and touch drag checks fail because real drags produce no service progress.
        & Real-input checks expose a core interaction failure despite other passing checks. \\
        \addlinespace
        \textit{Turbo Smash Beast}\newline DeepSeek-V4-Flash
        & Reset or scenario loading stops natural-time simulation; acceleration, coasting, and related real-input checks fail.
        & Checks combining scenario preparation and real input expose a failure in their interaction. \\
        \bottomrule
    \end{tabularx}
\end{table}

\textit{Real-input checks expose failures in player interaction.}
In the GPT-6-Astra implementation of \textit{Diner Dasher}, mouse and touch drags should advance the service workflow. The artifact passes L1 and L2 P0/P2 checks, but fails the L2 P1 checks for serving customers through real mouse and touch drags: neither interaction produces service progress. These core requirement checks test whether player actions produce the required gameplay outcome.

\textit{Scenario-based checks expose failures in combined operations.}
In the DeepSeek-V4-Flash/Claude Code implementation of \textit{Turbo Smash Beast}, resetting or loading a scenario through the evaluation interface suppresses natural-time simulation, leaving real driving inputs unable to advance the vehicle. The evaluator records failures of acceleration, coasting, and related real-input behavior. The generation traces explain the mismatch: self-tests through the evaluation interface explicitly advance time, while a successful driving test starts from a freshly loaded page. Neither exercises the failing combination of scenario preparation, real input, and natural time progression (Appendix~\ref{app:failure-traces}). Executing this combined sequence exposes an interaction failure that the separate self-tests miss.

These cases provide empirical support for the evaluator's design by showing how complementary checks identify concrete violations of task requirements. Real-input checks test player interactions, and scenario-based execution reveals inconsistencies between prepared states and subsequent gameplay.

\FloatBarrier

\section{Related Work}
\label{sec:related-work}

\subsection{From Code Generation to Complete Artifacts}
Autonomous software generation combines code synthesis with the ability to inspect, execute, and revise an artifact. The Codex model~\citep{chen2021codex} and CodeRL~\citep{le2022coderl} establish language-model synthesis and execution-guided learning, while SWE-Agent~\citep{yang2024sweagent}, AutoCodeRover~\citep{zhang2024autocoderover}, and ChatDev~\citep{qian2024chatdev} develop interfaces, program analysis, and role-based collaboration for longer development workflows. These systems motivate evaluating the artifact produced by the entire coding-agent loop, since the delivered artifact reflects both model capabilities and the surrounding development tools.

Evaluation spans increasingly broad artifact scopes, from bounded programs and repository changes to complete applications. APPS~\citep{hendrycks2021apps} and EvalPlus~\citep{liu2023evalplus} test bounded programs. At the repository level, SWE-bench~\citep{jimenez2024swebench} and FEA-Bench~\citep{li2025feabench} test changes to existing repositories. CodeFlowBench~\citep{wang2026codeflowbench} and KoCo-Bench~\citep{jiang2026kocobench} extend evaluation to reuse, dependencies, and domain knowledge. At the application level, benchmarks including WebGen-Bench~\citep{lu2025webgen}, E2EDev~\citep{liu2026e2edev}, RAL-Bench~\citep{pan2026ralbench}, and SaaSBench~\citep{ren2026saasbench} address broader integration requirements; Vision2Web~\citep{he2026vision2web} and VISTA~\citep{guo2026vista} also emphasize visual and interactive evidence. \sys shares this application-level perspective and concentrates on input, runtime state, rendering, and gameplay rules in a self-contained browser-native artifact.

Beyond the scope of the generated artifact, recent work also studies how to diagnose behavioral failures and express verification requirements. VideoVIBE~\citep{xu2026videovibe} evaluates fine-grained failure diagnosis from human-operated recordings of generated webpages, with source code as complementary context. FlowCheck~\citep{vir2026flowcheck} expresses user-visible information-flow constraints and compiles them into deterministic CodeQL analyses. This emphasis on diagnosing failures and making verification requirements explicit complements our focus on checking generated games against specified runtime requirements.

\subsection{Game Generation and Execution Feedback}
Game-generation systems address two complementary problems: coordinating construction and obtaining useful feedback. LLMGG~\citep{hu2024gamegeneration} studies joint rule-and-level synthesis; GameGPT~\citep{chen2023gamegpt}, ChatGE~\citep{hong2025chatge}, and AutoUE~\citep{yin2026autoue} organize development through specialized roles, conversation, or engine-grounded workflows. OpenGame~\citep{jiang2026opengame} combines agentic game development with execution-grounded post-training.

Execution feedback connects generated code to player-visible behavior. CreativeGame~\citep{ma2026creativegame} combines mechanic-guided planning with programmatic rewards and runtime validation; Play2Code~\citep{huang2026guiagents} alternates coding and GUI-agent playtesting; and ALIVE~\citep{zhang2026alive} turns automated play into learning signals. GameCWM distillation~\citep{serapio2026gamecwm} and The Verifier is the Curriculum~\citep{zhou2026verifiercurriculum} further use verification to support model training. These approaches use execution and verification feedback to improve generation or learning. \sys evaluates the resulting artifacts against requirements and an evaluation interface specification fixed before generation, and studies how tool access, nominal turn budget, reasoning effort, and harness choice relate to strict task success.

\subsection{Behavioral Evaluation of Generated Games}
Game benchmarks differ in both construction scope and the mechanism used to exercise an artifact. GameDevBench~\citep{chi2026gamedevbench} and GameEngineBench~\citep{la2026gameenginebench} evaluate scoped development within Godot and Unreal Engine projects. JamBench~\citep{sun2026jamer} includes theme-driven project generation and completion at several code granularities, while V-GameGym~\citep{zhang2026vgamegym} studies text-to-Pygame generation. PlayEval~\citep{peng2026playcoder}, PlaytestArena~\citep{huang2026guiagents}, Mage~\citep{liu2026mage}, and OpenGame-Bench~\citep{jiang2026opengame} assess interactive artifacts using combinations of play, runtime, structural, and visual evidence.

Related evaluators differ in how evaluation criteria are defined and how artifacts are exercised to obtain evidence. WebGameBench~\citep{zhang2026webgamebench} uses specification-guided browser interaction and permits candidate state preparation before the final user-level action. GameCraft-Bench~\citep{luo2026gamecraft} requires complete Godot projects and replayable demonstrations, including scenario initialization, and scores replay evidence against a hidden rubric. GameGen-Verifier~\citep{jia2026gamegenverifier} extracts precondition-interaction-postcondition keypoints and grounds them in each generated implementation through runtime state injection.
GameXpert-Bench~\citep{chen2026gamexpert} broadens evaluation to game generation, repair, and iterative refinement. Its generation track constructs shared event rubrics after generation by pooling events from completed artifacts and incorporating human review, then verifies them through code inspection and live interaction.

\sys fixes the evaluation interface specification and executable checks before generation. Task authors define scenario, action, and observation semantics, which agents implement alongside the game. This enables the same checks to evaluate artifacts with different internal implementations. Interface-based state observations, together with real-input and browser evidence, help assess whether the required behavior is reflected in actual gameplay.

\FloatBarrier

\section{Conclusion}
\label{sec:conclusion}

In this paper, we introduce \sys, which makes behavioral testability part of autonomous game generation through an evaluation interface specification declared before generation. Its 47 browser-native tasks combine source-level checks with browser-executed checks using semantic observations, real input, and runtime evidence. Across nine agent stacks, the highest observed mean L2 check pass rate is 93.2\%, yet the highest observed strict task success rate is only 55.3\%, exposing the gap between passing most checks and satisfying every required check. Comparisons of tool access, nominal turn budget, reasoning effort, and harness choice further reveal configuration-dependent outcomes. The benchmark provides a common protocol and task-level diagnostics for studying requirement compliance.

\FloatBarrier
\Needspace{7\baselineskip}
\bibliographystyle{unsrt}
\bibliography{references}

\clearpage
\appendix
\section{Supplementary Trace Analysis}
\label{app:trace-cases}

This appendix records selected diagnostic examples from evaluation reports and generation traces supporting the experimental discussion. The cases illustrate failure and repair mechanisms. Unless otherwise specified, DeepSeek-V4-Flash cases use the Claude Code baseline with full tool access, maximum reasoning effort, and a nominal turn budget of 120. Strict task success has the benchmark-compliance meaning defined in the main text.

\subsection{Runtime and Interaction Failures}
\label{app:failure-traces}

The following cases supplement the real-input and scenario-based diagnostics in Section~\ref{sec:threats}. They describe the prepared state, the interaction exercised, and the resulting check outcome.

\textbf{Real-input service progression.}
In \textit{Diner Dasher} generated by GPT-6-Astra with Codex CLI, two L2 P1 checks load \texttt{tray\_with\_correct\_item}, then obtain the tray item's and customer's screen bounds from the snapshot. One check sends a real mouse drag between their centers; the other sends the corresponding touch sequence. Each compares the resulting snapshot with the prepared state and requires an increase in service progress or earnings. Both fail at this progress assertion: neither real drag advances the service workflow. The checks also test tray consumption and visible rendering changes after successful service, but these runs fail before reaching those assertions. The artifact passes L1 and L2 P0/P2, showing why exposing a callable interface and a readable playfield does not establish that the required serving interaction works.

\textbf{Scenario preparation and natural-time progression.}
In the DeepSeek-V4-Flash/Claude Code implementation of \textit{Turbo Smash Beast}, reset or scenario loading enables test mode and suppresses natural-time simulation. The generation trace exercises two different paths: self-tests through the evaluation interface advance simulation time explicitly, while a successful real-input driving test starts from a freshly loaded page. The benchmark instead prepares a scenario and then exercises real driving input with natural time progression. In this sequence, the simulation remains stalled, and acceleration, coasting, and related real-input checks fail. The failing sequence combines operations that the separate self-tests do not exercise together.

\subsection{Execution-Guided Repairs}
\label{app:repair-traces}

These traces supplement the tool comparison in Section~\ref{sec:rq-tools} and the harness workflow analysis by connecting execution feedback to specific edits and subsequent checks.

\textbf{Feedback under different tool settings.}
For DeepSeek-V4-Flash's \textit{Diner Dasher}, the file read/write only configuration supports a correction of touch-release coordinates through source rereading; the final artifact passes the corresponding mouse and touch checks. In the run with full tool access, Chromium mouse input exposes a different defect: a hidden completion screen missed by stub-based tests. Correcting the screen-state mapping restores that transition. These runs illustrate the distinct repairs prompted by source inspection and browser interaction.

\textbf{Syntax repair and behavioral verification.}
In the \textit{Turbo Smash Beast} run with files + syntax checking, \texttt{node -{}-check} catches an unclosed construct. Closing it restores parsing, but acceleration and coasting still fail benchmark evaluation. The syntax feedback resolves a parsing defect while leaving the gameplay failures observable to behavioral checks.

\textbf{State-progression repair.}
In the \textit{Garden Gulp} run with full tool access, a browser-executed progression probe using the evaluation interface reveals that the hole's actual size never catches up with its target size. Repairing the update loop allows the probe to complete the level. This feedback links an evolving game-state discrepancy to a concrete update-loop repair.

\textbf{Browser-specific rendering repair.}
In DeepSeek-V4-Pro's \textit{Neon Flow}, browser execution exposes a negative-radius Canvas error missed by Node-based tests. The agent repairs the rendering path and verifies it again; the final artifact satisfies strict task success. This sequence illustrates feedback from executing the artifact with the browser's rendering API.

\textbf{Different repairs under two harnesses.}
Both maximum-effort DeepSeek-V4-Flash \textit{Sortie} runs achieve strict task success after execution-guided repairs. In Codex CLI, 27 self-tests through the evaluation interface pass before a real mouse drag fails. Inspection reveals that item positions are rendered relative to a layer below the heads-up display while the snapshot reports viewport coordinates; correcting this offset is followed by successful real-input retesting. In Claude Code, self-tests expose stale item states surviving reset because layout reconstruction preserves state indiscriminately. Restricting that preservation to resize is followed by a 148/148 self-test pass. Later edits repair the snap animation and adjust audio initialization before the final regression checks. The paired traces show how both harnesses support testing and repair, with different defects and follow-up edits in these runs. The self-test counts refer to different agent-authored tests.

\subsection{Nominal-Budget Trace Details}
\label{app:budget-cases}

Claude Code stops execution when the configured turn limit is reached. The following cases supplement Section~\ref{sec:rq-tools} by showing how this cutoff can interrupt verification or finalization before delivery is complete.

\textbf{Cutoff before verification executes.}
\textit{Ancient Beast} reaches the cutoff in both shorter-budget conditions. One 60-turn attempt stops immediately after writing a browser harness, before running it. The trace thus ends after preparing a verification step but before obtaining its execution feedback.

\textbf{Cutoff after self-tests pass.}
One 30-turn \textit{Barbie and Ken's Puzzles} attempt continues with screenshot processing after its self-tests pass, then reaches the cutoff without completing delivery. This case shows that passing the agent's self-tests does not itself complete the generation and delivery workflow.

\section{Generation and Evaluation Implementation Details}
\label{app:implementation}
\label{sec:system-boundary}

\textbf{Generation execution and delivery preflight.}
Each attempt uses the original task documents in a clean workspace and a fresh session. Claude Code runs in prompt mode with model and turn-budget arguments and stops when the configured turn limit is reached. Codex CLI uses an ephemeral session with workspace write access and JSON tracing. The host records execution traces, exit status, elapsed time, failure reasons, artifact status, and available usage measurements. The container launcher also monitors elapsed time and periods without trace updates to terminate timed-out runs. Evaluation requires a successful generation-process exit and a regular, non-symlink, nonempty \texttt{index.html} containing \texttt{</html>}. Artifacts that pass this preflight proceed to L1 and L2 evaluation.

\textbf{Task access and evaluation isolation.}
The launcher mounts the generation workspace with read/write access and marks the copied task documents read-only. The coding agent receives the generation prompt, gameplay design requirement, and evaluation interface specification (\texttt{tdd.md}); executable checks and evaluation runners are not mounted in the generation container. After generation ends, checks run in a separate evaluation container with the submitted HTML and tests mounted read-only.

\end{document}